\documentclass[nohyperref]{article}

\usepackage{microtype}
\usepackage{graphicx}
\usepackage{subfigure}
\usepackage{booktabs} % for professional tables

\usepackage{hyperref}

\usepackage[accepted]{icml2023}

\usepackage{amsmath}
\usepackage{amssymb}
\usepackage{mathtools}
\usepackage{amsthm}
\usepackage{placeins}
\usepackage[capitalize,noabbrev]{cleveref}

\theoremstyle{plain}
\newtheorem{theorem}{Theorem}[section]
\newtheorem{proposition}[theorem]{Proposition}

\theoremstyle{definition}

\newtheorem{assumption}[theorem]{Assumption}
\theoremstyle{remark}

\usepackage[textsize=tiny]{todonotes}

\usepackage{math_commands}

\usepackage{bbm}

\icmltitlerunning{Representation-Driven Reinforcement Learning}
\begin{document}

\twocolumn[
% \icmltitle{reprl: A Framework for Representation-Driven Reinforcement Learning}
\icmltitle{Representation-Driven Reinforcement Learning}


\icmlsetsymbol{equal}{*}

\begin{icmlauthorlist}
\icmlauthor{Ofir Nabati}{tech}
\icmlauthor{Guy Tennenholtz}{google}
\icmlauthor{Shie Mannor}{tech,nvidia}

\end{icmlauthorlist}

\icmlaffiliation{tech}{Department of Electrical-Engineering, Technion Institute of Technology, Israel}
\icmlaffiliation{nvidia}{Nvidia Research}
\icmlaffiliation{google}{Technion (currently at Google Research)}

\icmlcorrespondingauthor{Ofir Nabati}{ofirnabati@gmail.com}

\icmlkeywords{Machine Learning, ICML}

\vskip 0.3in
]



\printAffiliationsAndNotice{}  % leave blank if no need to mention equal contribution

\begin{abstract}
    We present a representation-driven framework for reinforcement learning. By representing policies as estimates of their expected values, we leverage techniques from contextual bandits to guide exploration and exploitation. Particularly, embedding a policy network into a linear feature space allows us to reframe the exploration-exploitation problem as a representation-exploitation problem, where good policy representations enable optimal exploration. We demonstrate the effectiveness of this framework through its application to evolutionary and policy gradient-based approaches, leading to significantly improved performance compared to traditional methods. Our framework provides a new perspective on reinforcement learning, highlighting the importance of policy representation in determining optimal exploration-exploitation strategies.
\end{abstract}

\section{Introduction}

Reinforcement learning (RL) is a field in machine learning in which an agent learns to maximize a reward through interactions with an environment. The agent maps its current state into action and receives a reward signal. Its goal is to maximize the cumulative sum of rewards over some predefined (possibly infinite) horizon \cite{sutton1998reinforcement}. This setting fits many real-world applications such as recommendation systems \cite{li2010contextual}, board games \cite{silver2017mastering}, computer games \cite{mnih2015human}, and robotics \cite{polydoros2017survey}. 

A large amount of contemporary research in RL focuses on gradient-based policy search methods \cite{sutton1999policy, dpg, trpo, schulman2017proximal, sac}. Nevertheless, these methods optimize the policy \textbf{locally} at specific states and actions. \citet{evolutionstr} have shown that such optimization methods may cause high variance updates in long horizon problems, while \citet{tessler2019distributional} have shown possible convergence to suboptimal solutions in continuous regimes. Moreover, policy search methods are commonly sample inefficient, particularly in hard exploration problems, as policy gradient methods usually converge to areas of high reward, without sacrificing exploration resources to achieve a far-reaching sparse reward. 

In this work, we present Representation-Driven Reinforcement Learning (RepRL) -- a new framework for policy-search methods, which utilizes theoretically optimal exploration strategies in a learned latent space. Particularly, we reduce the policy search problem to a contextual bandit problem, using a mapping from policy space to a linear feature space. Our approach leverages the learned linear space to optimally tradeoff exploration and exploitation using well-established algorithms from the contextual bandit literature \citep{abbasi2011improved, agrawal2013thompson}. By doing so, we reframe the exploration-exploitation problem to a representation-exploitation problem, for which good policy representations enable optimal exploration.

We demonstrate the effectiveness of our approach through its application to both evolutionary and policy gradient-based approaches -- demonstrating significantly improved performance compared to traditional methods. Empirical experiments on the MuJoCo \cite{MuJoCo} and MinAtar \cite{young19minatar} show the benefits of our approach, particularly in sparse reward settings. While our framework does not make the exploration problem necessarily easier, it provides a new perspective on reinforcement learning, shifting the focus to policy representation in the search for optimal exploration-exploitation strategies.

\section{Preliminaries}

We consider the infinite-horizon discounted Markov Decision Process (MDP). An MDP is defined by the tuple $\mathcal{M} = (\mathcal{S},\mathcal{A},r,T,\beta, \gamma)$, where $\mathcal{S}$ is the state space, $\mathcal{A}$ is the action space, $T: \mathcal{S} \times \mathcal{A} \rightarrow \Delta(\mathcal{S})$ is the transition kernel, $r: \mathcal{S} \times \mathcal{A} \rightarrow [0,1]$ is the reward function, $\beta \in \Delta(\mathcal{S})$ is the initial state distribution, and  $\gamma \in [0,1)$ is the discount factor. A stationary policy $\pi: \mathcal{S} \rightarrow \Delta(\mathcal{A})$, maps states into a distribution over actions. We denote by $\Pi$ the set of stationary stochastic policies, and the history of policies and trajectories up to episode $k$ by $\mathcal{H}_k$. Finally, we denote $S = \abs{\gS}$ and $A = \abs{\gA}$.

The return of a policy is a random variable defined as the discounted sum of rewards 
\begin{align}
G(\pi) = \sum_{t=0}^\infty \gamma^t r(s_t,a_t),
\label{def:return}
\end{align}
where $ s_0 \sim \beta, a_t \sim \pi(s_t), s_{t+1} \sim T(s_t,a_t)$, and the policy's value is its mean, i.e., $v(\pi) = \expect*{}{\sum_{t=0}^\infty \gamma^t r(s_t,a_t) | \beta , \pi,  T}$. An optimal policy maximizes the value, i.e.,
$
\pi^* \in \argmax_{\pi \in \Pi} v(\pi).
$

We similarly define the per-state value function, $v(\pi,s)$ as
$
    v(\pi,s) = \expect*{}{\sum_{t=0}^\infty \gamma^t r(s_t,a_t) | s_0=s , \pi,  T},
$
and note that $v(\pi) = \expect*{s \sim \beta}{v(\pi,s)}$.

Finally, we denote the discounted state-action frequency distribution w.r.t. $\pi$ by 
$$
  \rho^{\pi} (s,a) = (1-\gamma) \sum_{t=0}^{\infty} \gamma^t Pr\bigg(s_t=s, a_t=a | \beta , \pi, T\bigg),
$$
and let $\gK = \brk[c]*{\rho^\pi : \pi \in \Pi}$.

\subsection{Linear Bandits}
\label{section: linear bandits}

In this work, we consider the linear bandit framework as defined in \citet{abbasi2011improved}. At each time $t$, the learner is given a decision set $D_t \subseteq \sR^d$, which can be adversarially and adaptively chosen. The learner chooses an action $x_t \in D_t$ and receives a reward $r_t$, whose mean is linear w.r.t $x_t$, i.e., $\expect*{}{r_t | x_t} = \inner{x_t, w}$ for some unknown parameter vector $w \in \sR^d$.  

A general framework for solving the linear bandit problem is the ``Optimism in the Face of Uncertainty Linear bandit algorithm" (OFUL, \citet{abbasi2011improved}). There, a linear regression estimator is constructed each round as follows:
\begin{align}
    &\hat w_t = V_t^{-1} b_t, \nonumber \\
    &V_t = V_{t-1} + x_t x_t^\top,  \nonumber \\
    &b_t = b_{t-1} + x_t y_t, 
\label{bandit_update}
\end{align}
where $y_t, x_t$ are the noisy reward signal and chosen action at time $t$, respectively, and $V_0 = \lambda I$ for some positive parameter $\lambda > 0$. 

It can be shown that, under mild assumptions, and with high probability, the self-normalizing norm $\norm{\hat w_t - w}_{V_t}$ can be bounded from above \citep{abbasi2011improved}. OFUL then proceeds by taking an optimistic action $(x_t, \bar w_t) \in \arg\max_{x \in D_t, \bar w \in \gC_t} \inner{x, \bar w}$, where $\gC_t$ is a confidence set induced by the aforementioned bound on $\norm{\hat w_t - w}_{V_t}$. In practice, a softer version is used in \citet{chu2011contextual}, where an action is selected optimistically according to 
\begin{align*}
    x_t \in \arg\max_{x \in D_t} \inner{x, \hat w_t} + \alpha \sqrt{x^TV_t^{-1}x},
    \tag{OFUL}
\end{align*}
where $\alpha > 0$ controls the level of optimism.

Alternatively, linear Thompson sampling (TS, \citet{abeille2017linear}) shows it is possible to converge to an optimal solution with sublinear regret, even with a constant probability of optimism. This is achieved through the sampling of a parameter vector from a normal distribution, which is determined by the confidence set $\gC_t$. Specifically, linear TS selects an action according to
\begin{align*}
    x_t \in \arg\max_{x \in D_t} \inner{x, \tilde{w}_t},
    ~\tilde{w}_t \sim \gN\brk*{\hat w_t, \sigma^2 V_t^{-1}}
    \tag{TS},
\end{align*}
where $\sigma > 0$ controls the level of optimism. We note that for tight regret guarantees, both $\alpha$ and $\sigma$ need to be chosen to respect the confidence set $\gC_t$. Nevertheless, it has been shown that tuning these parameters can improve performance in real-world applications \citep{chu2011contextual}.

\section{RL as a Linear Bandit Problem}

\begin{figure}[t!]
\centering
\includegraphics[width=0.8\linewidth]{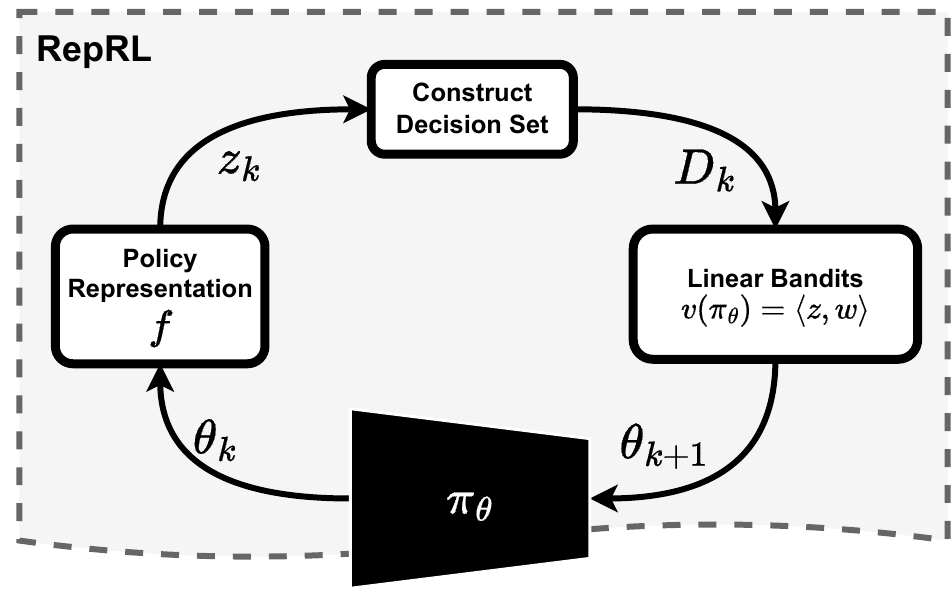}
 \caption{RepRL scheme. Composed of 4 stages: representation of the parameters, constructing a decision set, choosing the best arm using an off-the-shelf linear bandit algorithm, collect data with the chosen policy.}
 \label{fig:reprl}
\end{figure}

Classical methods for solving the RL problem attempted to use bandit formulations \citep{rlasbandit}. There, the set of policies $\Pi$ reflects the set of arms, and the value $v(\pi)$ is the expected bandit reward. Unfortunately, such a solution is usually intractable due to the exponential number of policies (i.e., bandit actions) in $\Pi$.

Alternatively, we consider a linear bandit formulation of the RL problem. Indeed, it is known that the value can be expressed in linear form as
\begin{align}
v(\pi)  
= 
\expect*{(s,a)\sim \rho^{\pi}}{r(s,a)} 
=
\inner{\rho^{\pi}, r}.
\label{linear_conection}
\end{align}
Here, any $\rho^\pi \in \gK$ represents a possible action in the linear bandit formulation \citep{abbasi2011improved}. Notice that $\abs{\gK} = \abs{\Pi}$, as any policy $\pi \in \Pi$ can be written as $\pi(a|s) = \frac{\rho^\pi(s,a)}{\sum_{a'} \rho^\pi(s, a')}$, rendering the problem intractable. Nevertheless, this formulation can be relaxed using a lower dimensional embedding of $\rho^\pi$ and $r$. As such, we make the following assumption.

\begin{assumption}[Linear Embedding] There exist a mapping $f: \Pi  \rightarrow \sR^d$ such that $v(\pi) =  \inner{f(\pi), w}$ for all $\pi \in \Pi$ and some unknown $w \in \sR^d$.
\label{assum1}
\end{assumption}

We note that \Cref{assum1} readily holds when $d = SA$ for $f(\pi) \equiv \rho^\pi$ and $w = r$. For efficient solutions, we consider environments for which the dimension $d$ is relatively low, i.e., $d \ll SA$.

Note that neural bandit approaches also consider linear representations \cite{riquelme2018deep}. Nevertheless, these methods use \textbf{mappings from states} $\gS \mapsto \sR^d$, whereas we consider \textbf{mapping entire policies} $\Pi \mapsto \sR^d$ (i.e., embedding the \emph{function} $\pi$). Learning a mapping $f$ can be viewed as trading the effort of finding good exploration strategies in deep RL problems to finding a good representation. We emphasize that we do not claim it to be an \emph{easier} task, but rather a \emph{different} viewpoint of the problem, for which possible new solutions can be derived. Similar to work on neural-bandits \cite{riquelme2018deep}, finding such a mapping requires alternating between representation learning and exploration. 

\subsection{RepRL}

We formalize a representation-driven framework for RL, inspired by linear bandits (\Cref{section: linear bandits}) and \Cref{assum1}. We parameterize the policy $\pi$ and mapping $f$ using neural networks, $\pi_\theta$ and $f_\phi$, respectively. Here, a policy $\pi_\theta$ is represented in lower-dimensional space as $f_\phi(\pi_\theta)$. Therefore, searching in policy space is equivalent to searching in the parameter space. With slight abuse of notation, we will denote $f_\phi(\pi_\theta) = f_\phi(\theta)$.

Pseudo code for RepRL is presented in \Cref{reprl_alg}. At every episode $k$, we map the policy's parameters $\theta_{k-1}$ to a latent space using $f_{\phi_{k-1}}(\theta_{k-1})$. We then use a construction algorithm, $\texttt{ConstructDecisonSet}(\theta_{k-1}, \mathcal{H}_{k-1})$, which takes into account the history $\mathcal{H}_{k-1}$, to generate a new decision set $D_k$. Then, to update the parameters $\theta_{k-1}$ of the policy, we select an optimistic policy $\pi_{\theta_k} \in D_k$ using a linear bandit method, such as TS or OFUL (see \Cref{section: linear bandits}). Finally, we rollout the policy $\pi_{\theta_k}$ and update the representation network and the bandit parameters according to the procedure outlined in \Cref{bandit_update}, where $x_k$ are the learned representations of $f_{\phi_k}$. A visual schematic of our framework is depicted in Figure \ref{fig:reprl}.

In the following sections, we present and discuss methods for representation learning, decision set construction, and propose two implementations of RepRL in the context of evolutionary strategies and policy gradient. We note that RepRL is a framework for addressing RL through representation, and as such, any representation learning technique or decision set algorithm can be incorporated as long as the basic structure is maintained.

\begin{algorithm}[t!]
    \caption{RepRL}
    \begin{algorithmic}[1]
    \STATE \textbf{Init:} $\mathcal{H}_0 \gets \emptyset$, $\pi_{\theta_0}$, $f_{\phi_0}$ randomly initialized
    \FOR{ $k = 1,2,\ldots$ } 
        \STATE {\color{gray}\emph{Representation Stage:}}\\ Map the policy network $\pi_{\theta_{k-1}}$ using representation network $f_{\phi_{k-1}}(\theta_{k-1})$.
        \STATE {\color{gray}\emph{Decision Set Stage:}}\\ $D_k \gets \texttt{ConstructDecisonSet}(\theta_{k-1}, \mathcal{H}_{k-1})$.
        \STATE {\color{gray}\emph{Bandit Stage:}}\\ Use linear bandit algorithm to choose $\pi_{\theta_k}$ out of $D_k$.
        \STATE {\color{gray}\emph{Exploitation Stage:}}\\ Rollout policy $\pi_{\theta_k}$ and store the return $G_k$ in $\gH_k$. \\~\\
        \STATE Update representation $f_{\phi_k}$.
        \STATE Update bandit parameters $\hat w_t, V_t$ (\Cref{bandit_update}) with the updated representation.
    
    \ENDFOR
    \end{algorithmic}
    \label{reprl_alg}
    \end{algorithm}

\subsection{Learning Representations for RepRL}

We learn a linear representation of a policy using tools from variational inference. Specifically, we sample a representation from a posterior distribution $z \sim f_\phi(z | \theta)$, and train the representation by maximizing the Evidence Lower Bound (ELBO) \citep{kingma2013auto}
$
        \mathcal{L}(\phi, \kappa) = -\mathbb{E}_{z \sim f_\phi(z | \theta)} \brk[s]*{\log p_\kappa(G|z)}
        +D_{KL}(f_
    \phi(z| \theta) \| p(z)),
$
where $f_\phi(z | \theta)$ acts as the encoder of the embedding, and $p_\kappa(G|z)$ is the return decoder or likelihood term. 

\begin{figure}[t!]
\centering
\includegraphics[width=0.7\linewidth]{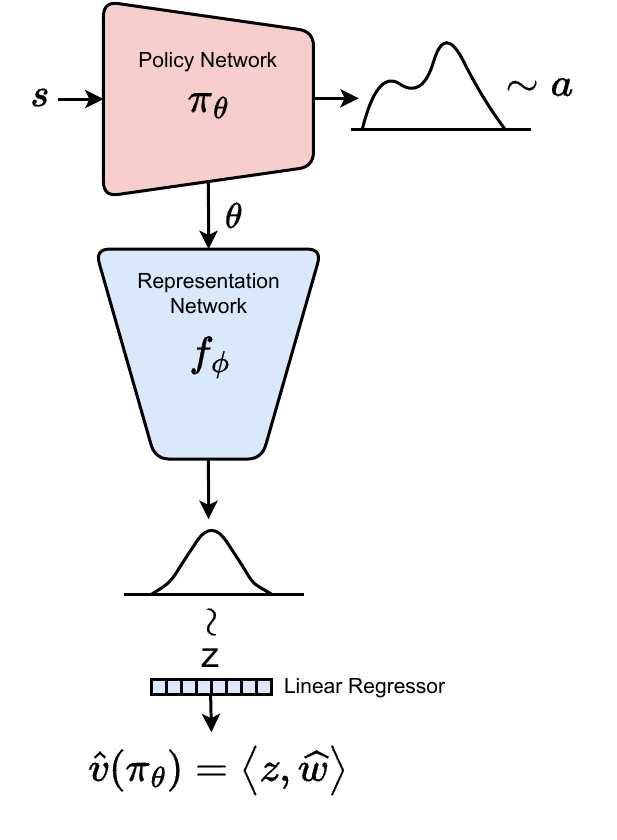}
 \caption{The diagram illustrates the structure of the networks in RepRL. The policy's parameters are fed into the representation network, which acts as a posterior distribution for the policy's latent representation. Sampling from this posterior, the latent representation is used by the bandits algorithm to evaluate the value that encapsulates the exploration-exploitation tradeoff.}
  \label{fig:networks}
\end{figure}

The latent representation prior $p(z)$ is typically chosen to be a zero-mean Gaussian distribution. In order to encourage linearity of the value (i.e the return's mean) with respect to the learned representation, we chose the likelihood to be a Gaussian distribution with a mean that is linear in the representation, i.e., $p_\kappa(G|z) = \mathcal{N}(\kappa^\top z, \sigma^2)$. When the encoder is also chosen to be a Gaussian distribution, the loss function has a closed form. The choice of the decoder to be linear is crucial, due to the fact that the value is supposed to be linear w.r.t learned embeddings.  
The parameters $\phi$ and $\kappa$ are the learned parameters of the encoder and decoder, respectively. Note that a deterministic mapping occurs when the function $f_\phi(z|\theta)$ takes the form of the Dirac delta function. A schematic of the architectural framework is presented in \cref{fig:networks}.

\subsection{Constructing a Decision Set}
\label{sec:decision_set}

The choice of the decision set algorithm (line 4 of \Cref{reprl_alg}) may have a great impact on the algorithm in terms of performance and computational complexity. Clearly, choosing $D_k = \Pi, \forall k$ will be unfeasible in terms of computational complexity. Moreover, it may be impractical to learn a linear representation for all policies at once. We present several possible choices of decision sets below.

\paragraph{Policy Space Decision Set. }
One potential strategy is to sample a set of policies centered around the current policy
\begin{align}
D_k = \{\theta_k + \epsilon_i\}_{i=1}^N, \; \; \epsilon_i \sim \mathcal{N}(0, \nu^2 I),
\label{eq: policy space decision set}
\end{align}
where $\nu>0$ controls how local policy search is. This approach is motivated by the assumption that the representation of policies in the vicinity of the current policy will exhibit linear behavior with respect to the value function due to their similarity to policies encountered by the learner thus far. 

\paragraph{Latent Space Decision Set. } An alternative approach involves sampling policies in their learned latent space, i.e.,
\begin{align}
D_k = \{z_k + \epsilon_i\}_{i=1}^N, \; \; \epsilon_i \sim \mathcal{N}(0, \nu^2 I),
\label{eq: latent space decision set}
\end{align}
where $z_k \sim f_\phi(z|\theta_k)$. The linearity of the latent space ensures that this decision set will improve the linear bandit target (UCB or the sampled value in TS), which will subsequently lead to an improvement in the actual value.  This approach enables optimal exploration w.r.t. linear bandits, as it uniformly samples the eigen directions of the precision matrix $V_t$, rather than only sampling specific directions as may occur when sampling in the parameter space.

Unlike \Cref{eq: policy space decision set} constructing the set in \Cref{eq: latent space decision set} presents several challenges. First, in order to rollout the policy $\pi_{\theta_k}$, one must construct an inverse mapping to extract the chosen policy from the selected latent representation. This can be done by training a decoder for the policy parameters $q(\theta|z)$. Alternatively, we propose to use a decoder-free approach. Given a target embedding  $z^* \in \argmax_{z \in D_t} \inner{z, \hat w}$, we search for a policy 
$
\theta^* \in \argmax_\theta f_\phi(z^*|\theta).
$
This optimization problem can be solved using gradient descent-based optimization algorithms by varying the inputs to $f_\phi$. 
A second challenge for latent-based decision sets involves the realizability of such policies. That is, there may exist representations $z \in D_k$, which are not mapped by any policy in $\Pi$. Lastly, even for realizable policies, the restored $\theta$ may be too far from the learned data manifold, leading to an overestimation of its value and a degradation of the overall optimization process. One way to address these issues is to use a small enough value of $\nu$ during the sampling process, reducing the probability of the set members being outside the data distribution. We leave more sophisticated methods of latent-based decision sets for future work.

\paragraph{History-based Decision Set.}
An additional approach uses the history of policies at time $k$ to design a decision set. Specifically, at time episode $k$ we sample around the set of policies observed so far, i.e.,
\begin{align}
D_k 
= 
\bigcup_{\ell \in [k]} \{\theta_\ell + \epsilon_{\ell,i}\}_{i=1}^N, \; \; \epsilon_{\ell,i} \sim \mathcal{N}(0, \nu^2 I),
\label{eq: history space decision set}
\end{align}
resulting in a decision set of size $Nk$. After improving the representation over time, it may be possible to find a better policy near policies that have already been used and were missed due to poor representation or sampling mismatch. This method is quite general, as the history can be truncated only to consider a certain number of past time steps, rather than the complete set of policies observed so far. Truncating the history can help reduce the size of the decision set, making the search more computationally tractable.

In \Cref{section: experiments}, we compare the various choices of decision sets. Nevertheless, we found that using policy space decisions is a good first choice, due to their simplicity, which leads to stable implementations. Further exploration of other decision sets is left as a topic for future research.

\subsection{Inner trajectory sampling}
\label{sec:inner trajectory sampling}

Vanilla RepRL uses the return values of the entire trajectory. As a result, sampling the trajectories at their initial states is the natural solution for both the bandit update and representation learning. However, the discount factor diminishes learning signals beyond the $\frac{1}{1 - \gamma}$ effective horizon, preventing the algorithm from utilizing these signals, which may be critical in environments with long-term dependencies. On the other hand, using a discount factor $\gamma = 1$ would result in returns with a large variance, leading to poor learning. 
Instead of sampling from the initial state, we propose to use the discount factor and sample trajectories at various states during learning, enabling the learner to observe data from different locations along the trajectory. Under this sampling scheme, the estimated value would be an estimate of the following quantity:
$$
\tilde v(\pi) = \expect*{s \sim \rho^\pi}{v(\pi,s)}.
$$
In the following proposition we prove that optimizing $\tilde v(\pi)$ is equivalent to optimizing the real value.

\begin{proposition}
For a policy $\pi \in \Pi$,  and stationary $\rho^\pi$ we get
$
\tilde v(\pi) = \frac{v(\pi)}{1 - \gamma}.
$
\label{prop_avg_reward}
\end{proposition}
The proof can be found in the \cref{appendix:proof}. That is, sampling along the trajectory from $\rho^\pi$ approximates the scaled value, which, like $v(\pi)$, exhibits linear behavior with respect to the reward function. Thus, instead of sampling the return defined in \cref{def:return}, we sample
$
\tilde G(\pi) = \sum_{t=0}^\infty \gamma^t r(s_t,a_t),
$
where $ s_0 \sim \rho^\pi, a_t \sim \pi(s_t), s_{t+1} \sim T(s_t,a_t)$, both during representation learning and bandit updates. Empirical evidence suggests that uniformly sampling from the stored trajectory produces satisfactory results in practice. 
% As for the bandit parameter updates, we also sample the return from a random $t$ along the trajectory, while using only single trajectory for the update due to the 

\begin{algorithm}[t!]
\caption{Representation Driven Evolution Strategy}
\begin{algorithmic}[1]
\STATE \textbf{Input:} initial policy $\pi_0=\pi_{\theta_0}$, noise $\nu$, step size $\alpha$,  decision set size $N$, history $\mathcal{H}$.
\FOR{ $t = 1,2,\ldots,T$ }   
    \STATE Sample an evaluation set and collect their returns.
    \STATE Update representation $f_t$ and bandit parameters $(\hat w_t, V_t)$ using history. 
    \STATE Construct a decision set $D_t$.
    \STATE Use linear bandit algorithm to evaluate each policy in $D_t$.
    \STATE Update policy using ES scheme (\cref{sec:repes}).
\ENDFOR
\end{algorithmic}
\label{alg_reprl_es}
\end{algorithm}

\section{RepRL Algorithms}

In this section we describe two possible approaches for applying the RepRL framework; namely, in Evolution Strategy \cite{wierstra2014natural} and Policy Gradients \cite{sutton1999policy}.

\subsection{Representation Driven Evolution Strategy}
\label{sec:repes}
Evolutionary Strategies (ES) are used to train agents by searching through the parameter space of their policy and sampling their return. In contrast to traditional gradient-based methods, ES uses a population of candidates evolving over time through genetic operators to find the optimal parameters for the agent. Such methods have been shown to be effective in training deep RL agents in high-dimensional environments \cite{evolutionstr, ars}.

At each round, the decision set is chosen over the policy space with Gaussian sampling around the current policy as described in \cref{sec:decision_set}. 
\Cref{alg_reprl_es} considers an ES implementation of RepRL. To improve the stability of the optimization process, we employ soft-weighted updates across the decision set. This type of update rule is similar to that used in ES algorithms \cite{evolutionstr, ars}, and allows for an optimal exploration-exploitation trade-off, replacing the true sampled returns with the bandit's value. Moreover, instead of sampling the chosen policy, we evaluate it by also sampling around it as done in ES-based algorithms. Each evaluation is used for the bandit parameters update and representation learning process. Sampling the evaluated policies around the chosen policy helps the representation avoid overfitting to a specific policy and generalize better for unseen policies - an important property when selecting the next policy.

\begin{algorithm}[t!]
\caption{Representation Driven Policy Gradient}
\begin{algorithmic}[1]
\STATE \textbf{Input:} initial policy $\pi_\theta$, decision set size $N$,  history $\mathcal{H}$.
\FOR{ $t = 1,2,\ldots,T$ } 
    \STATE Collect trajectories using $\pi_\theta$.
    \STATE Update representation $f$ and bandit parameters $(\hat w_t, V_t)$ using history. 
    \STATE Compute Policy Gradient loss  $\mathcal{L}_{PG}(\theta)$.
    \STATE Sample a decision set and choose the best policy $\tilde \theta$. 
    \STATE Compute gradient of the regularized Policy Gradient loss with $d(\theta,\tilde \theta)$ (\Cref{eq: reg pg loss}).

\ENDFOR
\end{algorithmic}
\label{reprl_policy_gradient}
\end{algorithm}

Unlike traditional ES, optimizing the UCB in the case of OFUL or sampling using TS can encourage the algorithm to explore unseen policies in the parameter space. This exploration is further stabilized by averaging over the sampled directions, rather than assigning the best policy in the decision set. This is particularly useful when the representation is still noisy,  reducing the risk of instability caused by hard assignments. An alternative approach uses a subset of $D_t$ with the highest bandit scores, as suggested in \citet{ars}, which biases the numerical gradient towards the direction with  the highest potential return.    

\begin{figure*}
\centering
\includegraphics[width=0.35\linewidth]{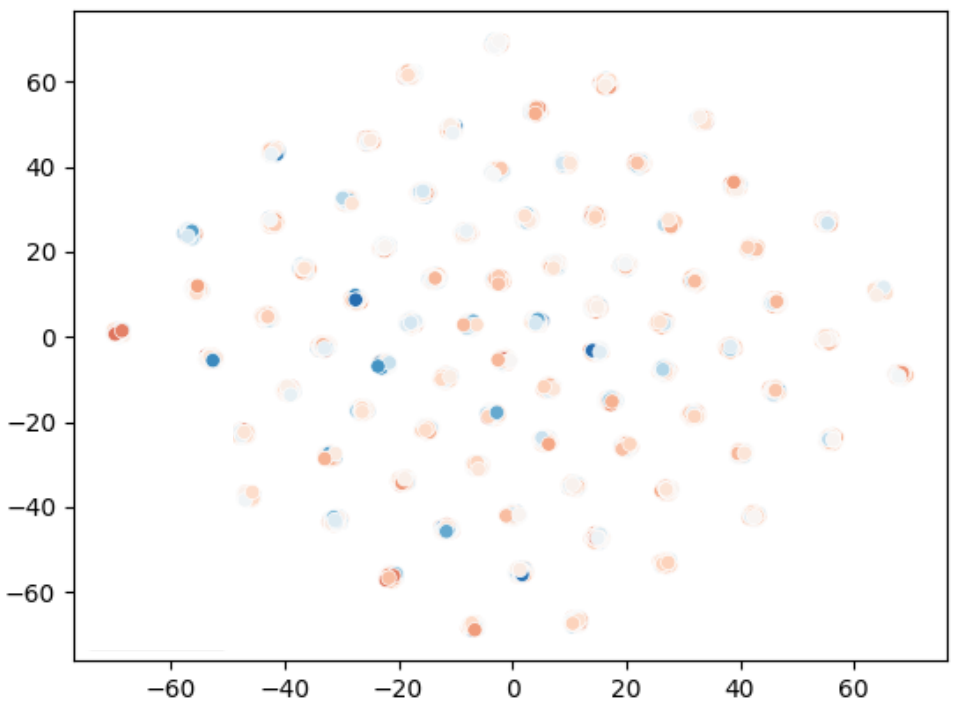}
\includegraphics[width=0.35\linewidth]{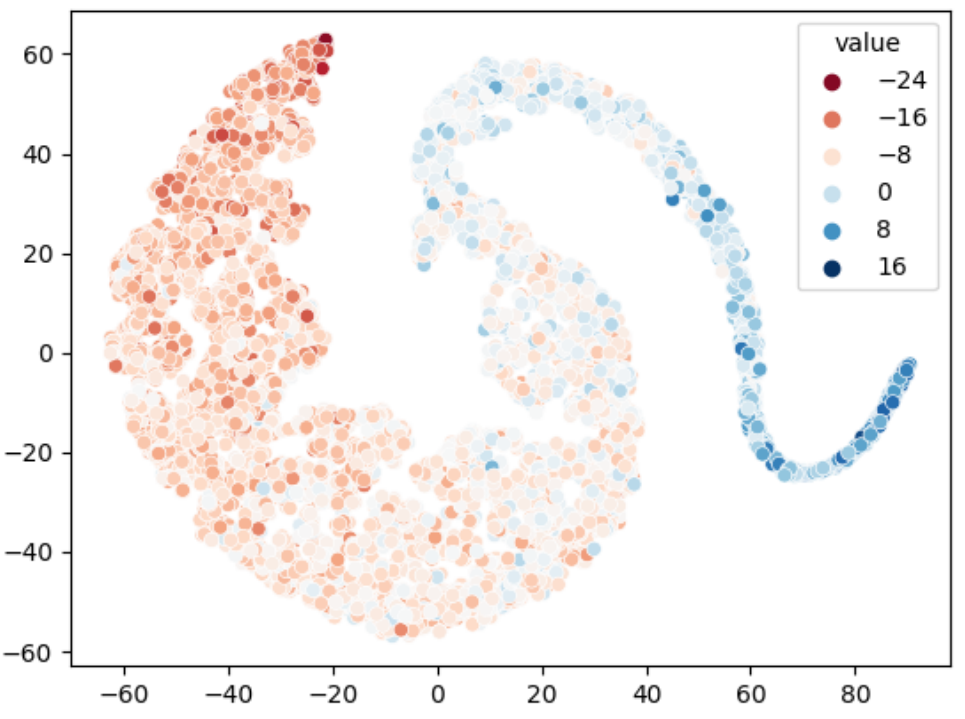}
\caption{The two-dimensional t-SNE visualization depicts the policy representation in the GridWorld experiment. On the right, we observe the learned latent representation, while on the left, we see the direct representation of the policy's weights. Each point in the visualization corresponds to a distinct policy, and the color of each point corresponds to a sample of the policy's value.}    
  \label{fig:gridworld_tsne}
\end{figure*}

\begin{figure}[t!]
\centering
\includegraphics[width=1.0\linewidth]{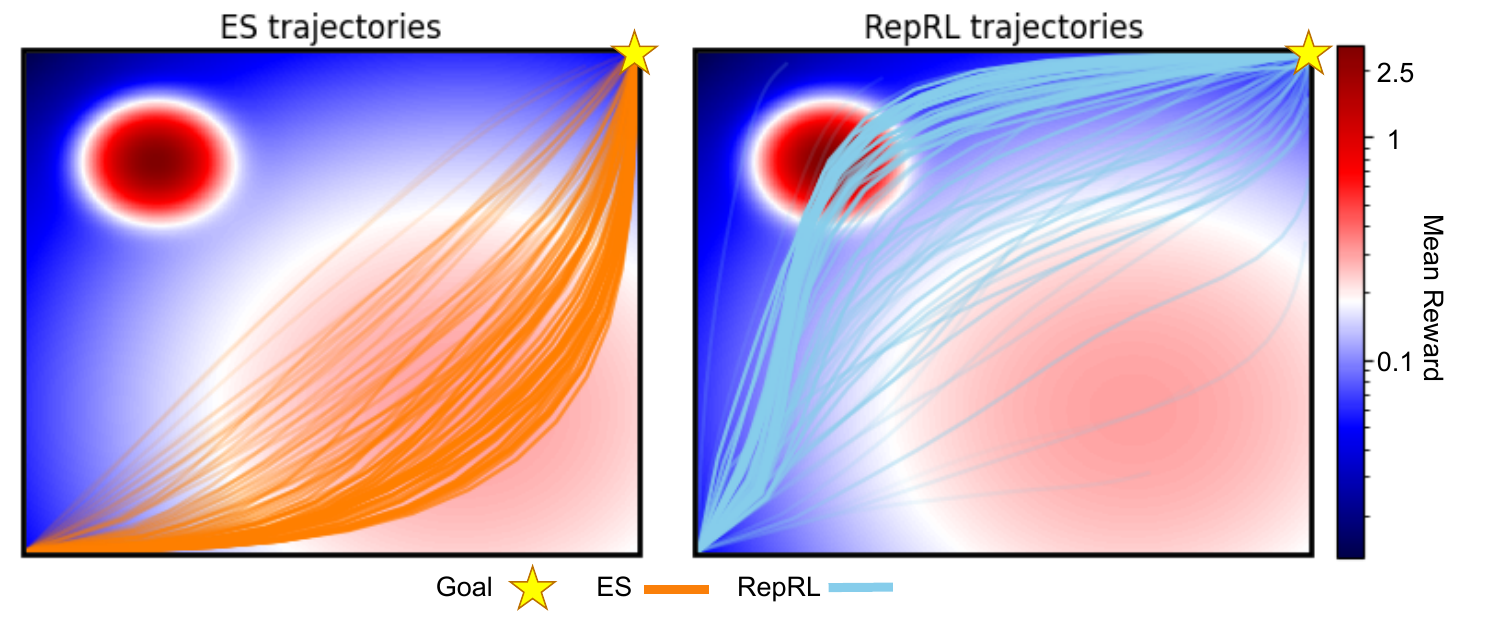}
  \caption{GridWorld visualization experiment. Trajectories were averaged across 100 seeds at various times during training, where more recent trajectories have greater opacity. Background colors indicate the level of mean reward.} 
  \label{fig:gridworld}
\end{figure}

\subsection{Representation Driven Policy Gradient}
RepRL can also be utilized as a regularizer for policy gradient algorithms. Pseudo code for using RepRL in policy gradients is shown in \cref{reprl_policy_gradient}. At each gradient step, a weighted regularization term $d(\theta,\tilde\theta)$ is added, where $\tilde \theta$ are the parameters output by RepRL with respect to the current parameters for a chosen metric (e.g., $\ell_2$):
\begin{align}
\label{eq: reg pg loss}
    \mathcal{L}_{\text{reg}}(\theta) = \mathcal{L}_{\text{PG}}(\theta) + \zeta d(\theta,\tilde \theta).
\end{align}

After collecting data with the chosen policy and updating the representation and bandit parameters, the regularization term is added to the loss of the policy gradient at each gradient step. The policy gradient algorithm can be either on-policy or off-policy while in our work we experiment with an on-policy algorithm.  

Similar to the soft update rule in ES, using RepRL as a regularizer can significantly stabilize the representation process. Applying the regularization term biases the policy toward an optimal exploration strategy in policy space. This can be particularly useful when the representation is still weak and the optimization process is unstable, as it helps guide the update toward more promising areas of the parameter space. In our experiments, we found that using RepRL as a regularizer for policy gradients improved the stability and convergence of the optimization process.

\section{Experiments}
\label{section: experiments} 

In order to evaluate the performance of RepRL, we conducted experiments on various tasks on the MuJoCo \cite{MuJoCo} and MinAtar \cite{young19minatar} domains. We also used a sparse version of the MuJoCo environments, where exploration is crucial. We used linear TS as our linear bandits algorithm as it exhibited good performance during evaluation. The detailed network architecture and hyperparameters utilized in the experiments are provided in \cref{appendix:hpyerparameters}.

\begin{figure*}[t!]
\centering
\includegraphics[width=0.95\linewidth]{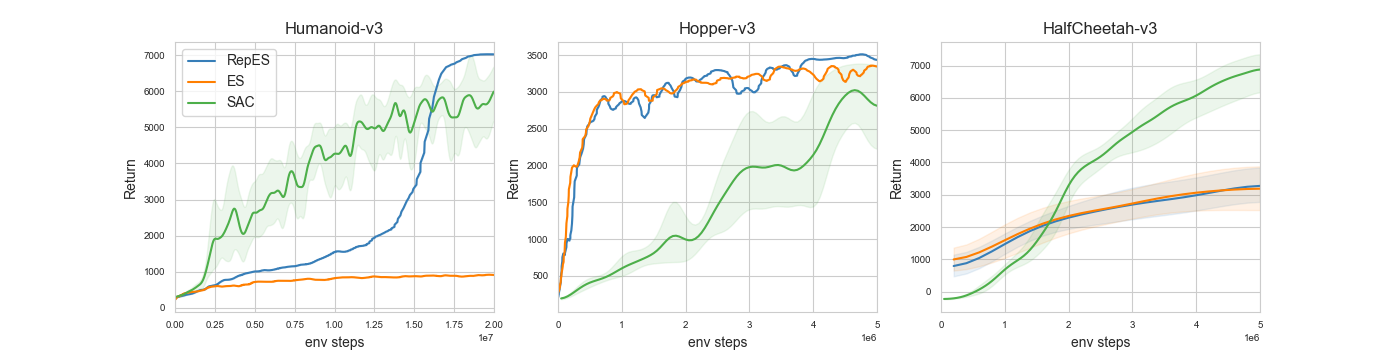}
\includegraphics[width=0.95\linewidth]{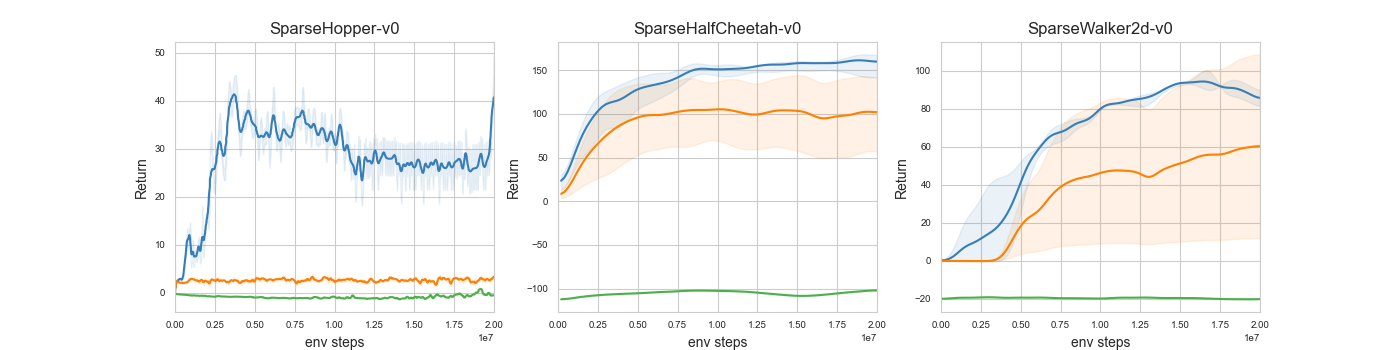}
\caption{MuJoCo experiments during training. The results are for the MuJoCo suitcase (top) and the modified sparse MuJoCo (bottom).}    
 \label{fig:MuJoCo}
\end{figure*}

\paragraph{Grid-World Visualization.}

Before presenting our results, we demonstrate the RepRL framework on a toy example. Specifically, we constructed a GridWorld environment (depicted in \Cref{fig:gridworld}) which consists of spatially changing, noisy rewards. The agent, initialized at the bottom left state $(x,y) = (1,1)$, can choose to take one of four actions: up, down, left, or right. To focus on exploration, the rewards were distributed unevenly across the grid. Particularly, the reward for every $(x,y)$ was defined by the Normal random variable
$
    r(x,y) \sim \mathcal{N}\brk*{\mu(x,y), \sigma^2},
$
where $\sigma >0$ and
$
    \mu(x,y) 
    \propto 
    R_1\exp\brk[c]*{-\frac{(x-x_1)^2 + (y-y_1)^2}{a_1}} 
    +
    R_2\exp\brk[c]*{-\frac{(x-x_2)^2 + (y-y_2)^2}{a_2}} 
    +
    R_3 \mathbbm{1}_{\brk[c]*{(x,y) = \text{goal}}}.
$
That is, the reward consisted of Normally distributed noise, with mean defined by two spatial Gaussians, as shown in \Cref{fig:gridworld}, with $R_1 > R_2$, $a_1 < a_2$ and a goal state (depicted as a star), with $R_3 \gg R_1, R_2$. Importantly, the values of $R_1, R_2, R_3, a_1, a_2$ were chosen such that an optimal policy would take the upper root in \Cref{fig:gridworld}.

Comparing the behavior of RepRL and ES on the GridWorld environment, we found that RepRL explored the environment more efficiently, locating the optimal path to the goal. This emphasizes the varying characteristics of state-space-driven exploration vs. policy-space-driven exploration, which, in our framework, coincides with representation-driven exploration. \cref{fig:gridworld_tsne} illustrates a two-dimensional t-SNE plot comparing the learned latent representation of the policy with the direct representation of the policy weights.

\paragraph{Decision Set Comparison.}
We begin by evaluating the impact of the decision set on the performance of the RepRL. For this, we tested the three decision sets outlined in \Cref{sec:decision_set}. The evaluation was conducted using the Representation Driven Evolution Strategy variant on a sparse HalfCheetah environment. A history window of 20 policies was utilized when evaluating the history-based decision set. A gradient descent algorithm was employed to obtain the parameters that correspond to the selected latent code in the latent-based setting

As depicted in \Cref{fig:decison_sets} at \cref{appendix:decision_set}, RepRL demonstrated similar performance for the varying decision sets on the tested domains. In what follows, we focus on policy space decision sets.

\paragraph{MuJoCo.}
We conducted experiments on the MuJoCo suitcase task using RepRL. Our approach followed the setting of \citet{ars}, in which a linear policy was used and demonstrated excellent performance on MuJoCo tasks. We utilized the ES variant of our algorithm (\cref{alg_reprl_es}). We incorporated a weighted update between the gradients using the bandit value and the zero-order gradient of the sampled returns, taking advantage of sampled information and ensuring stable updates in areas where the representation is weak. 

We first evaluated RepES on the standard MuJoCo baseline (see \cref{fig:MuJoCo}). RepES either significantly outperformed or performed on-par with ES. We also tested a modified, sparse variant of MuJoCo. In the sparse environment, a reward was given for reaching a goal each distance interval, denoted as $d$, where the reward function was defined as:
$$
r(s,a) = 
\begin{cases}
10 - c(a), & |x_{\text{agent}}| \bmod d = 0 \\
-c(a), & \text{o.w.}
\end{cases}
$$
Here, $c(a)$ is the control cost associated with utilizing action $a$, and $x_{agent}$ denotes the location of the agent along the $x$-axis. The presence of a control cost function incentivized the agent to maintain its position rather than actively exploring the environment. The results of this experiment, as depicted in \cref{fig:MuJoCo}, indicate that the RepRL algorithm outperformed both the ES and SAC algorithms in terms of achieving distant goals. However, it should be noted that the random search component of the ES algorithm occasionally resulted in successful goal attainment, albeit at a significantly lower rate in comparison to the RepRL algorithm.

\begin{figure*}[t!]
\centering
\includegraphics[width=0.85\linewidth]{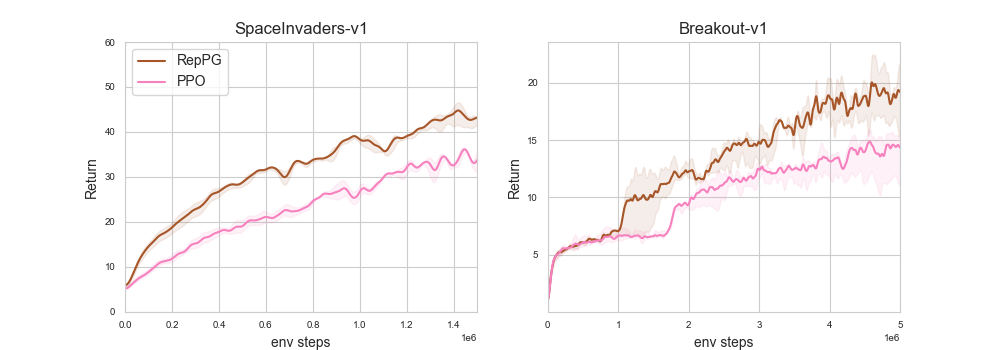}
 \caption{MinAtar experiments during training.}
  \label{fig:minatar}
\end{figure*}

\paragraph{MinAtar.}
We compared the performance of RepRL on MinAtar \citep{young19minatar} with the widely used policy gradient algorithm PPO \cite{schulman2017proximal}. Specifically, we compared PPO against its regularized version with RepRL, as described in \Cref{reprl_policy_gradient}, and refer to it as RepPG. We parametrized the policy by a neural network. 
Although PPO collects chunks of rollouts (i.e., uses subtrajectories), RepPG adjusted naturally due to the inner trajectory sampling (see \Cref{sec:inner trajectory sampling}). That is, the critic was used to estimate the value of the rest of the trajectory in cases where the rollouts were truncated by the algorithm.   

Results are shown in \cref{fig:minatar}. Overall, RepRL outperforms PPO on all tasks, suggesting that RepRL is effective at solving challenging tasks with sparse rewards, such as those found in MinAtar.

% \subsection{Abalation}
% TBD

\section{Related Work}

\textbf{Policy Optimization:} Policy gradient methods \cite{sutton1999policy} have shown great success at various challenging tasks, with numerous improvements over the years; most notable are policy gradient methods for deterministic policies  \cite{dpg,ddpg}, trust region based algorithms \cite{trpo,schulman2017proximal}, and maximum entropy algorithms \cite{sac}. 
Despite its popularity, traditional policy gradient methods are limited in continuous action spaces. Therefore, \citet{tessler2019distributional} suggest optimizing the policy over the policy distribution space rather than the action space.

In recent years, finite difference gradient methods have been rediscovered by the RL community. This class of algorithms uses numerical gradient estimation by sampling random directions \cite{nesterov2017random}. A closely related family of optimization methods is Evolution Strategies (ES) a class of black-box optimization algorithms that heuristic search by perturbing and evaluating the set members, choosing only the mutations with the highest scores until convergence. \citet{evolutionstr} used ES for RL as a zero-order gradient estimator for the policy, parameterized as a neural network. 
ES is robust to the choice of the reward function or the horizon length and it also does not need value function approximation as most state-of-art algorithms. Nevertheless, it suffers from low sample efficiency due to the potentially noisy returns and the usage of the final return value as the sole learning signal. Moreover, it is not effective in hard exploration tasks. \citet{ars} improves ES by using only the most promising directions for gradient estimation.

%\citet{pi-ars} improved the scalability of ES to more complex models by combining self-supervised learning for the first layers. In terms of injecting exploration, \citet{ns-es} used a heuristic novelty term in order to encourage exploring new parameters.          

\textbf{Policy Search with Bandits.} \citet{rlasbandit} was one of the first works to utilize multi-arm bandits for policy search over a countable stationary policy set -- a core approach for follow-up work \cite{burnetas1997optimal, agrawal1988asymptotically}. Nevertheless, the concept was left aside due to its difficulty to scale up with large environments. 

As an alternative, Neural linear bandits \cite{riquelme2018deep, neural_linear_ucb, nabati2021online} simultaneously train a neural network policy, while interacting with the environment, using a chosen linear bandit method and are closely related to the neural-bandits literature \cite{neuralucb, kassraie2022neural}. In contrast to this line of work, our work maps entire policy functions into linear space, where linear bandit approaches can take effect. This induces an exploration strategy in policy space, as opposed to locally, in action space.

\textbf{Representation Learning.} Learning a compact and useful representation of states \citep{laskin2020curl,schwartz2019language,tennenholtzuncertainty}, actions \citep{tennenholtz2019natural,chandak2019learning}, rewards \citep{barreto2017successor,nair2018visual,toro2019learning}, and policies \citep{hausman2018learning,eysenbach2018diversity}, has been at the core of a vast array of research. Such representations can be used to improve agents' performance by utilizing the structure of an environment more efficiently. 
Policy representation has been the focus of recent studies, including the work by \citet{tang2022inputting}, which, similar to our approach, utilizes policy representation to learn a generalized value function. They demonstrate that the generalized value function can generalize across policies and improve value estimation for actor-critic algorithms, given certain conditions. In another study, \citet{li2022erl} enhance the stability and efficiency of Evolutionary Reinforcement Learning (ERL) \cite{khadka2018evolution} by adopting a linear policy representation with a shared state representation between the evolution and RL components.
In our research, we view the representation problem as an alternative solution to the exploration-exploitation problem in RL. Although this shift does not necessarily simplify the problem, it transfers the challenge to a different domain, offering opportunities for the development of new methods.

\section{Discussion and Future Work}
We presented RepRL, a novel representation-driven framework for reinforcement learning. By optimizing the policy over a learned representation, we leveraged techniques from the contextual bandit literature to guide exploration and exploitation. We demonstrated the effectiveness of this framework through its application to evolutionary and policy gradient-based approaches, leading to significantly improved performance compared to traditional methods. 

In this work, we suggested reframing the exploration-exploitation problem as a representation-exploitation problem. By embedding the policy network into a linear feature space, good policy representations enable optimal exploration. This framework provides a new perspective on reinforcement learning, highlighting the importance of policy representation in determining optimal exploration-exploitation strategies.

As future work, one can incorporate RepRL into more involved representation methods, including pretrained large Transformers \cite{devlin2018bert, brown2020language}, which have shown great promise recently in various areas of machine learning. Another avenue for future research is the use of RepRL in scenarios where the policy is optimized in latent space using an inverse mapping (i.e., decoder), as well as more involved decision sets. Finally, while this work focused on linear bandit algorithms, future work may explore the use of general contextual bandit algorithms, (e.g., SquareCB \citet{foster2020beyond}), which are not restricted to linear representations.

\section{Acknowledgments}
This work was partially funded by the Israel Science Foundation under Contract 2199/20.

\bibliography{iclr2023_conference}
\bibliographystyle{iclr2023_conference}

%%%%%%%%%%%%%%%%%%%%%%%% וSUPP FROM HERE ##################

\onecolumn
\icmltitle{Representation-Driven Reinforcement Learning - Appendix}
\appendix
\section{Algorithms}

\begin{algorithm}
\caption{Random Search / Evolution Strategy}
\begin{algorithmic}[1]
\STATE \textbf{Input:} initial policy $\pi_0=\pi_{\theta_0}$, noise $\nu$, step size $\alpha$, set size $K$.  
\FOR{ $t = 1,2,\ldots,T$ }   
    \STATE Sample a decision set $D_t = \{\theta_{t-1} \pm \delta_i \}_{i=1}^K$, $\delta_i \sim \mathcal{N}(0, \nu^2 I)$.
    \STATE Collect the returns $\{G(\theta_{t-1} \pm \delta_i)\}_{i=1}^K$ of each policy in $D_t$.
    \STATE Update policy 
    $$
    \theta_t = \theta_{t-1} +\frac{\alpha}{\sigma_R K} \sum_{i=1}^K \bigg[G(\theta_{t-1} + \delta_i) - G(\theta_{t-1} - \delta_i) \bigg]\delta_i   
    $$
    
\ENDFOR
\end{algorithmic}
\label{alg_random_search}
\end{algorithm}

\begin{algorithm}
\caption{Representation Driven Evolution Strategy}
\begin{algorithmic}[1]
\STATE \textbf{Input:} initial policy $\pi_0=\pi_{\theta_0}$, noise $\nu$, step size $\alpha$, set size $K$, decision set size $N$, history $\mathcal{H}$.
\FOR{ $t = 1,2,\ldots,T$ }   
    \STATE Sample an evaluation set $\{\theta_{t-1} \pm \delta_i \}_{i=1}^K$, $\delta_i \sim \mathcal{N}(0, \nu^2 I)$.
    \STATE Collect the returns $\{G(\theta_{t-1} \pm \delta_{i})\}_{i=1}^K$ from the environment and store them in replay buffer.
    \STATE Update representation $f_t$ and bandit parameters $(\hat w_t, V_t)$ using history. 
    \STATE Construct a decision set $D_t = \{\theta_{t-1} \pm \delta_i \}_{i=1}^N$, $\delta_i \sim \mathcal{N}(0, \nu^2 I)$.
    \STATE Use linear bandit algorithm to evaluate each policy in $D_t$: $\{\hat{v}(\theta_{t-1} \pm \delta_i)\}_{i=1}^N$ .

    \STATE Update policy 
    \begin{align*}
           g_t =& \frac{1}{N} \sum_{i=1}^N \bigg[\hat v(\theta_{t-1} + \delta_{i}) - \hat v(\theta_{t-1} - \delta_{i}) \bigg]\delta_i,\\
           \theta_t = & \theta_{t-1}  +\alpha g_t
    \end{align*}
\ENDFOR
\end{algorithmic}
\label{alg_reprl_es}
\end{algorithm}

\begin{algorithm}
\caption{Representation Driven Policy Gradient}
\begin{algorithmic}[1]
\STATE \textbf{Input:} initial policy $\pi_\theta$, noise $\nu$, step size $\alpha$, decision set size $N$, $\zeta$, history $\mathcal{H}$.
\FOR{ $t = 1,2,\ldots,T$ } 
    \FOR{ $1,2,\ldots,K$ }   
        \STATE Collect trajectory data using $\pi_\theta$.
        \STATE Update representation $f$ and bandit parameters $(\hat w, \Sigma)$ using history. 
    \ENDFOR
    \FOR{ $1,2,\ldots,M$ }     
        \STATE Sample a decision set $D = \{\theta + \delta_i \}_{i=1}^N$, $\delta_i \sim \mathcal{N}(0, \nu^2 I)$.
        \STATE Use linear bandit algorithm to choose the best parameter $\tilde \theta \in \argmax_{\theta \in D} \inner{z, \hat w}$ for $z \sim f(z|\theta)$.
        \STATE Compute 
        \begin{align*}
               g =& \nabla_\theta \big[ \mathcal{L}_{PG}(\theta) + \zeta \| \theta - \tilde \theta\|_2 \big],\\
               \theta = & \theta  - \alpha g
        \end{align*}
    \ENDFOR
\ENDFOR
\end{algorithmic}
\label{reprl_policy_gradient}
\end{algorithm}

\FloatBarrier

\section{Variational Interface}
\label{appendix:vae}
We present here proof of the ELBO loss for our variational interface, which was used to train the representation encoder. 
\begin{proof}
\begin{align*}
    \log p(G;\phi,\kappa) &= \log \int_z p_\kappa(G|z) p(z) dz \\
    &= \log \int_z p_\kappa(G|z) \frac{p(z)}{f_
    \phi(z|\pi)} f_\phi(z|\pi) dz \\
    &= \log \mathbb{E}_{z \sim f_\phi(z|\pi)} \bigg[ p_\kappa(G|z) \frac{p(z)}{f_
    \phi(z|\pi)} \bigg] \\
    &\geq \mathbb{E}_{z \sim f_\phi(z|\pi)} \bigg[ \log p_\kappa(G|z) \bigg] + \mathbb{E}_{z \sim f_\phi(z|\pi)} \bigg[ \log \frac{p(z)}{f_
    \phi(z|\pi)} \bigg] \\
    &= \mathbb{E}_{z \sim f_\phi(z|\pi)}\bigg[ \log p_\kappa(G|z) \bigg] -D_{KL} (f_
    \phi(z|\pi) \| p(z)),
\end{align*}

where the inequality is due to Jensen's inequality.  

% \section{Proof for \cref{prop_avg_reward}}
\section{Proof for \cref{prop_avg_reward}}
\label{appendix:proof}
For stationary occupancy measure $\rho^\pi$ we get:
$$
\rho^\pi(s') = \sum_{s,a} \rho^\pi(s) \pi(a|s) T(s'|s,a) 
$$
By definition:
\begin{align*}
   \tilde v(\pi) &=  \sum_s \rho^\pi(s) v(\pi,s)\\
                 &=  \sum_s \rho^\pi(s) \sum_a \pi(a|s) \brk[c]*{r(s,a) + \gamma \sum_{s'} T(s'|s,a) v(\pi,s')}\\
                 &=  v(\pi) + \gamma \sum_s \rho^\pi(s) \sum_a \pi(a|s) \sum_{s'} T(s'|s,a) v(\pi,s')\\
                 &=  v(\pi) + \gamma \sum_{s'} \rho^\pi(s') v(\pi,s') \\
                 & = v(\pi) + \gamma \tilde v(\pi),
\end{align*}
    where the second equality is due to the Bellman equation and the third is from the definition. 
    Therefore, 
    $$
    \tilde v(\pi) = v(\pi) + \gamma \tilde v(\pi)  \Longrightarrow  \tilde v(\pi) = \frac{v(\pi)}{1 - \gamma}
    $$
\end{proof}

\section{Full RepRL Scheme}
The diagram presented below illustrates the networks employed in RepRL. The policy's parameters are inputted into the representation network, which serves as a posterior distribution capturing the latent representation of the policy. Subsequently, a sampling procedure is performed from the representation posterior, followed by the utilization of a linear return encoder, acting as the likelihood, to forecast the return distribution with a linear mean (i.e. the policy's value). This framework is employed to maximize the Evidence Lower Bound (ELBO).

\begin{figure*}[h]
\centering
\includegraphics[width=0.5\linewidth]{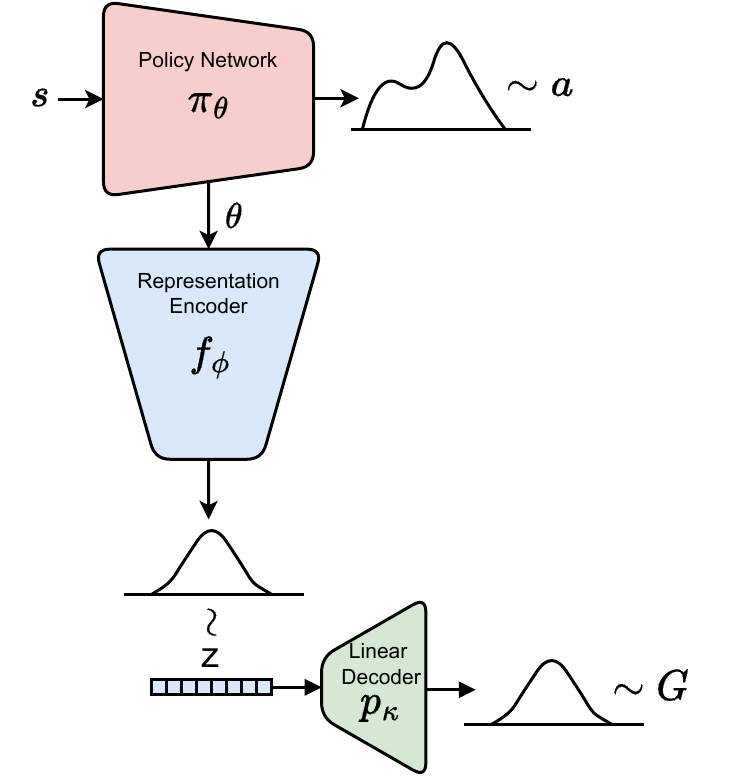}
 \caption{The full diagram illustrates the networks in RepRL.}
  \label{fig:mujoco}
\end{figure*}

\newpage
\section{Decision Set Experiment}
The impact of different decision sets on the performance of RepRL was assessed in our evaluation. We conducted tests using three specific decision sets as described in \cref{sec:decision_set}. The evaluation was carried out on a sparse HalfCheetah environment, utilizing the RepES variant. When evaluating the history-based decision set, we considered a history window consisting of $20$ policies. In the latent-based setting, the parameters corresponding to the selected latent code were obtained using a gradient descent algorithm.
The results showed that RepRL exhibited similar performance across the various decision sets tested in different domains.

\label{appendix:decision_set}
\begin{figure*}[h]
\centering
\includegraphics[width=0.5\linewidth]{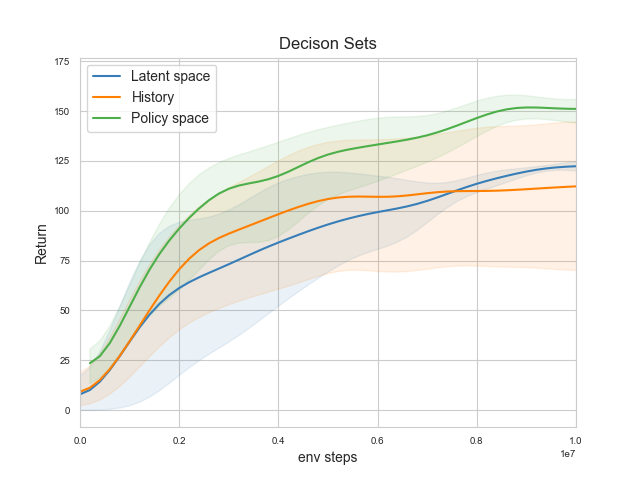}
\caption{Plots depict experiments for three decision sets: policy space-based, latent space-based, and history-based. The experiment was conducted on the SparseHalfCheetah environment.} 
 \label{fig:decison_sets}
\end{figure*}

\FloatBarrier
\section{Hyperparameters and Network Architecture}
\label{appendix:hpyerparameters}
\subsection{Grid-World}
\label{appendix:gridworld}
In the GridWorld environment, a $8 \times 8$ grid is utilized with a horizon of $20$, where the reward is determined by a stochastic function as outlined in the paper:
$
    r(x,y) \sim \mathcal{N}\brk*{\mu(x,y), \sigma^2},
$
where $\sigma >0$ and
$
    \mu(x,y) 
    \propto 
    R_1\exp\brk[c]*{-\frac{(x-x_1)^2 + (y-y_1)^2}{a_1}} 
    +
    R_2\exp\brk[c]*{-\frac{(x-x_2)^2 + (y-y_2)^2}{a_2}} 
    +
    R_3 \mathbbm{1}_{\brk[c]*{(x,y) = \text{goal}}}.
$
The parameters of the environment are set as $R_1=2.5, R_2=0.3, R_3 = 13, \sigma = 3, a_1 = 0.125, a_2=8$.

The policy employed in this study is a fully-connected network with 3 layers, featuring the use of the $tanh$ non-linearity operator. The hidden layers' dimensions across the network are fixed at $32$, followed by a $Softmax$ operation. The state is represented as a one-hot vector. please rephrase the next paragraph so it will sounds more professional:
The representation encoder is built from Deep Weight-Space (DWS) layers \cite{dwsnet}, which are equivariant to the permutation symmetry of fully connected networks and enable much stronger representation capacity of deep neural networks compared to standard architectures. The DWS model (DWSNet) comprises four layers with a hidden dimension of 16. Batch normalization is applied between these layers, and a subsequent fully connected layer follows. Notably, the encoder is deterministic, meaning it represents a delta function. For more details, we refer the reader to the code provided in \citet{dwsnet}, which was used by us. 
%with a representation encoder utilized to output the parameter vector, i.e. in the form of a delta function $f(z|\theta) = \delta(z-\theta)$.

In the experimental phase, $300$ rounds were executed, with $100$ trajectories sampled at each round utilizing noisy sampling of the current policy, with a zero-mean Gaussian noise and a standard deviation of $0.1$. The ES algorithm utilized a step size of $0.1$, while the RepRL algorithm employed a decision set of size $2048$ without a discount factor ($\gamma=1$) and $\lambda=0.1$.

\subsection{MuJoCo}
In the MuJoCo experiments, both ES and RepES employed a linear policy, in accordance with the recommendations outlined in \cite{ars}. For each environment, ES utilized the parameters specified by \citet{ars}, while RepES employed the same sampling strategy in order to ensure a fair comparison.

RepES utilized a representation encoder consisting of $4$ layers of a fully-connected network, with dimensions of $2048$ across all layers, and utilizing the $ReLU$ non-linearity operator. This was followed by a fully-connected layer for the mean and variance. The latent dimension was also chosen to be $2048$. After each sampling round, the representation framework (encoder and decoder) were trained for $3$ iterations on each example, utilizing an $Adam$ optimizer and a learning rate of $3e-4$. When combining learning signals of the ES with RepES, a mixture gradient approach was employed, with 20\% of the gradient taken from the ES gradient and 80\% taken from the RepES gradient. Across all experiments, a discount factor of $\gamma = 0.995$ and $\lambda=0.1$ were used.

\subsection{MinAtar}
In the MinAtar experiments, we employed a policy model consisting of a fully-connected neural network similar to the one utilized in the GridWorld experiment, featuring a hidden dimension of $64$. The value function was also of a similar structure, with a scalar output.  The algorithms collected five rollout chunks of $512$ between each training phase.

The regulation coefficient chosen for RepRL was $1$, while the discount factor and the mixing factor were set as $\gamma = 0.995$ and $\lambda=0.1$. The representation encoder used was similar to the one employed in the GridWorld experiments with two layers, followed by a symmetry invariant layer and two fully connected layers.

\end{document}